\documentclass[letterpaper, 10 pt, conference]{IEEEconf}  % Comment this line out if you need a4paper

\IEEEoverridecommandlockouts                              % This command is only needed if 
\usepackage{graphicx} % for pdf, bitmapped graphics files
\usepackage[ruled,vlined]{algorithm2e}
\usepackage[caption=false,font=footnotesize]{subfig}
\usepackage{bm}

\usepackage{siunitx}
\usepackage{tikz}
\usepackage{collcell}
\usepackage{colortbl}
\usepackage{hhline}
\usepackage{makecell}
\usepackage{multirow} 
\usepackage{ifthen}
\usepackage{xstring}
\usepackage{array}
\usepackage{amsmath}
\usepackage{amsfonts}
\usepackage{booktabs}
\usepackage{rotating} 

\newcolumntype{?}{!{\vrule width 1.5pt}}

\newcommand*{\MinNumber}{0}%
\newcommand*{\MaxNumber}{1}%
\newcommand{\ApplyGradient}[1]{%
        \StrMid{#1}{1}{4}[\tcr]
        \StrMid{#1}{6}{9}[\std]
        \StrMid{#1}{1}{10}[\tot]
        \StrMid{#1}{11}{11}[\isBold]
        \pgfmathsetmacro{\PercentColor}{100.0*(\tcr-\MinNumber)/(\MaxNumber-\MinNumber)}

        \edef\HeatCell{\noexpand\cellcolor{gray!\PercentColor!white}}%
        
        \newcommand{\CellValue}{#1}
        
        \IfDecimal{\tcr}{
            \ifthenelse{\equal{\isBold}{b}}{
                \renewcommand{\CellValue}{\textbf{\tcr} $\pm \std$}}
            {
                \renewcommand{\CellValue}{\tcr \ $\pm \std$}
            }
        }{}
        
        \HeatCell \CellValue
}
 
\newcolumntype{R}{>{\collectcell\ApplyGradient}m{1.6cm}<{\centering\endcollectcell}}
\newcolumntype{T}{>{\collectcell\ApplyGradient}m{1.04cm}<{\centering\endcollectcell}}
\newcolumntype{Q}{>{\collectcell}m{0.9cm}<{\endcollectcell}}
\newcolumntype{B}{>{\centering}m{1.8cm}}
\newcolumntype{C}{>{\centering}m{1.2cm}}

\title{\LARGE \bf
Zero-Shot Terrain Generalization for Visual Locomotion Policies}

\author{Alejandro~Escontrela$^{1, 2}$, George~Yu$^1$, Peng~Xu$^1$, Atil~Iscen$^1$, Jie~Tan$^1$ % <-this % stops a space
\thanks{$^1$ Google Brain Robotics,}
\thanks{{\tt \{georgeyu,pengxu,atil,jietan\}@google.com}}
\thanks{$^2$ Georgia Institute of Technology, {\tt aescontrela@gatech.edu}}
\thanks{Work performed while Alejandro was an intern at Google Brain.}}

\begin{document}

\maketitle
\thispagestyle{empty}
\pagestyle{empty}

% Abstract.
\begin{abstract}
Legged robots have unparalleled mobility on unstructured terrains. However, it remains an open challenge to design locomotion controllers that can operate in a large variety of environments. In this paper, we address this challenge of automatically learning locomotion controllers that can generalize to a diverse collection of terrains often encountered in the real world. We frame this challenge as a multi-task reinforcement learning problem and define each task as a type of terrain that the robot needs to traverse. We propose an end-to-end learning approach that makes direct use of the raw exteroceptive inputs gathered from a simulated 3D LiDAR sensor, thus circumventing the need for ground-truth heightmaps or preprocessing of perception information. As a result, the learned controller demonstrates excellent zero-shot generalization capabilities and can navigate 13 different environments, including stairs, rugged land, cluttered offices, and indoor spaces with humans.
\end{abstract}

\section{INTRODUCTION}
The ability to traverse unstructured terrains make legged robots an appealing solution to a wide variety of tasks, including disaster relief, last-mile delivery, industrial inspection, and planetary exploration \cite{albee20ieeeaero_mp_planetary_exploration, Bellicoso18jfr_advances_legged_robots}. To deploy robots in these settings successfully, we must design controllers that work well across many different terrains. Due to the diversity of environments that a legged robot can operate in, hand-engineering such a controller presents unique challenges. Deep Reinforcement Learning (DRL) has proven itself capable of automatically acquiring control policies to accomplish a large variety of challenging locomotion tasks. However, many of these approaches learn control policies that succeed in a single type of terrain with limited variations. This approach limits the robot's ability to generalize to new or unseen environments, which is a crucial feature of a useful locomotion controller. 

%Additionally, ground-truth heightmaps of the surrounding environment are often required as inputs to many approaches in the visual locomotion literature. Heightmaps are not directly observable and require sophisticated elevation mapping techniques which are prone to error \cite{Fankhauser2014CLAWAR}. Furthermore, visual locomotion policies must produce smooth and realizable actuation commands if they are to be successfully deployed in the real world. A significant drawback of reactive locomotion policies is that they learn jerky motion patterns which cannot be successfully executed on real hardware without leading to loss of balance or damage to the robot.

\begin{figure}[ht]
\centering
\subfloat[Office with humans in motion]{\includegraphics[width=0.48\linewidth]{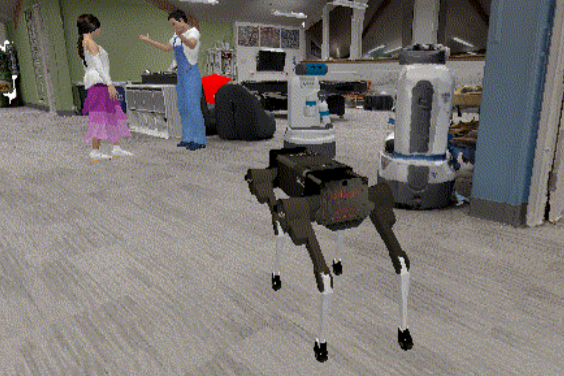}%
\label{fig:dynamic_env}}
\hfil
\subfloat[Office 1]{\includegraphics[width=0.48\linewidth]{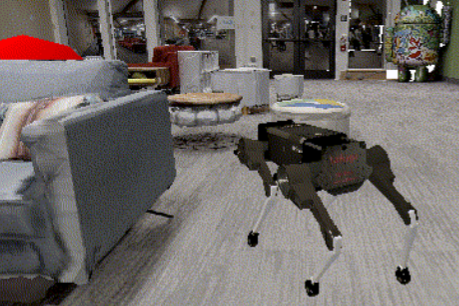}%
\label{fig:office_1}}
\hfil
\subfloat[Office 2]{\includegraphics[width=0.48\linewidth]{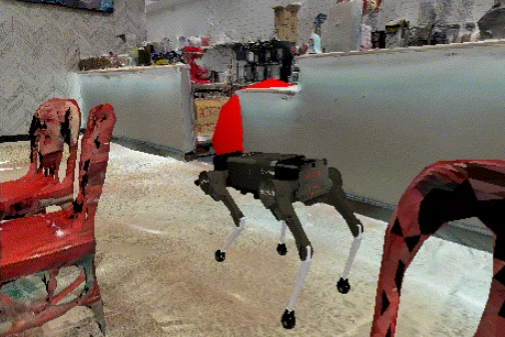}%
\label{fig:office_2}}
\hfil
\subfloat[Hilly]{\includegraphics[width=0.48\linewidth]{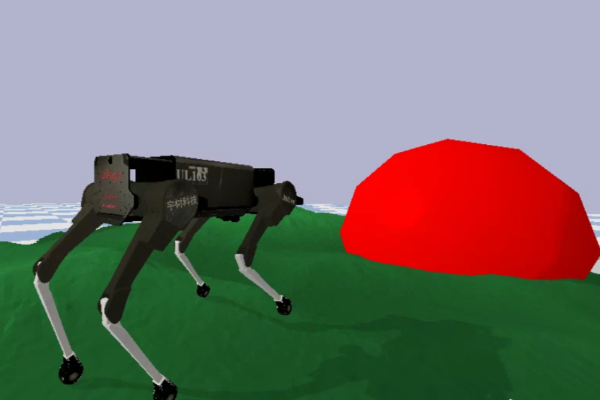}%
\label{fig:continuous}}
\hfil
\subfloat[Mountainous]{\includegraphics[width=0.48\linewidth]{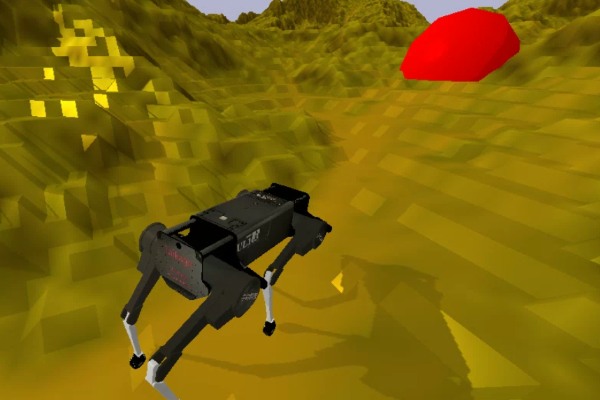}%
\label{fig:mountainous}}
\hfil
\subfloat[Maze]{\includegraphics[width=0.48\linewidth]{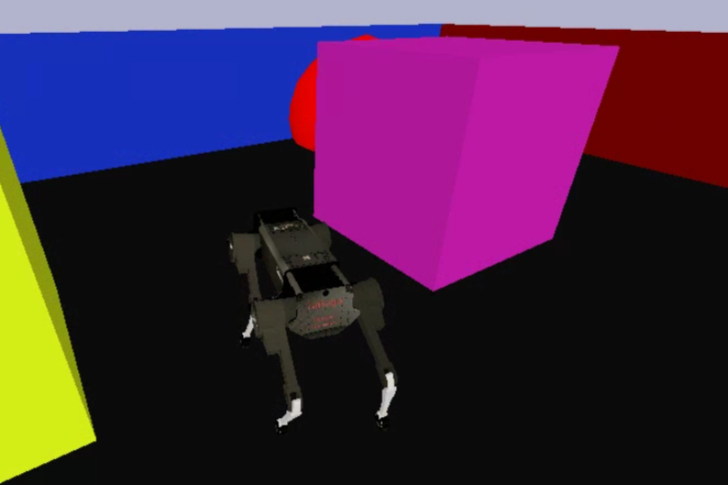}%
\label{fig:maze}}
\caption{A Laikago robot navigating a variety of complex terrains not encountered during training.}
\label{fig:eval_envs}
\end{figure} % Eval environments.

In this paper, we develop an end-to-end reinforcement learning system that enables legged robots to traverse a large variety of terrains. To facilitate learning generalizable policies, we make two purposeful design decisions for our learning system. First, we formulate the problem as a Multi-Task Partially Observable Markov Decision Problem and show that the robot learns a robust policy that works well across a wide variety of tasks (terrains). To this end, we develop a novel procedural terrain generation method, which can efficiently generate a large variety of terrains for training. Second, we design an end-to-end neural network architecture that can handle both perception and locomotion. We call this parameterization a \emph{visual-locomotion} policy.  While many prior works in the legged robot literature focused on blind walking, which does not involve exteroceptive sensors (e.g., camera, LiDAR), we find that exteroceptive perception is essential for robots to navigate in diverse environments. Our end-to-end visual-locomotion policy takes both exteroceptive (a LiDAR scan) and proprioceptive information of the robot and outputs low-level motor commands. We embed the Policies Modulating Trajectory Generator (PMTG) \cite{Iscen18pmlr_PMTG} framework into our policy architecture to generate cyclic and smooth actuation patterns, and to facilitate the learning of robust locomotion policies. 

We evaluate our learning system using a high-fidelity physics simulator \cite{coumans2019} and visually-realistic indoor scans \cite{fei18cvpr_gibson} (Figure 1). We test the learned policy in thirteen different and realistic simulation environments (five training and eight testing). Our system learns highly generalizable locomotion policies, which demonstrate zero-shot generalization to unseen testing environments. We also show that our visual-locomotion policy's parameterization is key to generalization and yields far better performance than commonly-used reactive policies. This paper's main contributions include an end-to-end visual-locomotion policy parameterization and a complete multi-task learning system, with which a quadruped robot learns a single locomotion policy that can traverse a diverse set of terrains.

%Additionally, we incorporate perception into our system by providing the agent with observations from a simulated 3D LiDAR sensor. We circumvent the need for manual preprocessing of the sensor observations (e.g., point cloud segmentation, heightmap generation, etc.) by utilizing the raw 3D LiDAR scan. 

\section{RELATED WORK}
\subsection{Legged Locomotion}
%Researchers have demonstrated that a variety of locomotion behaviours, ranging from stable walking to agile bounding, can be achieved through application of control techniques such as trajectory optimization \cite{winkler2018arxiv}, whole-body control \cite{kim16ieee}, model predictive control with Differential Dynamic Programming \cite{Grandia19}, and state-machines \cite{Bledt18cheetah}. These approaches often require expert knowledge of the robot's accurate dynamics and the resulted controller does not generalize well to multiple type of terrain. In this work, we demonstrate that a learned visual locomotion policy, which takes introspective and 3D LiDAR measurement as input, is able to control the legged robots to go through various type of terrain types in physics simulator. The result policy is also capable of zero-shot generalization to new terrains. All these are done without knowing either dynamics of robot or detailed geometry of the underlying terrain.

Locomotion controllers can be developed using trajectory optimization \cite{winkler2018arxiv}, whole-body control \cite{kim16ieee}, model predictive control \cite{Grandia19}, and state-machines \cite{Bledt18cheetah}. While the controllers developed by these techniques can generalize to a certain degree, expertise and manual tuning are often needed to adapt them to different terrains. 

In contrast, Deep Reinforcement Learning \cite{sutton2018reinforcement} can automatically learn agile and robust locomotion skills \cite{tan2018simtoreal, haarnoja2018learning, hwangbo2019learning, Leeeabc5986}. Prior work in RL has learned policies that are specific for a single environment \cite{ha2020learning}, or generalize to variations of a single type of terrain \cite{Hess17arxiv_locomotion_emergence, Peng17siggraph_deeploco,Tsounis2020arxiv_deepgait}. Recently, Lee \emph{et. al.} \cite{Leeeabc5986} combined various techniques, such as ActuatorNet \cite{hwangbo2019learning}, PMTG \cite{Iscen18pmlr_PMTG}, curriculum learning and ``learning by cheating'' \cite{chen2019learning}, which successfully performed zero-shot transfer from simulation to many challenging terrains in the real world. While our paper's high-level goal is similar to this prior work, our approach incorporates exteroceptive sensors that enable the robot to navigate in cluttered indoor environments where blind walking may have difficulties.

\subsection{Multi-Task Reinforcement Learning}
Multi-task reinforcement learning (MTRL) \cite{Caruana93icml_mtl} is a promising approach to train generalizable policies that can accomplish a wide variety of tasks. Hessel \emph{et. al.} \cite{Hessel18arxiv_multitask} learned a single policy that achieves state-of-the-art performance on 57 Atari games. Yu \emph{et al.} \cite{Yu19arxiv_meta_world} evaluated the performance of various RL algorithms on a grasping and manipulation benchmark and demonstrated that a single control policy is capable of completing a variety of complex robotic manipulation tasks. In this paper, we apply MTRL to develop a learning system for locomotion that enables legged robots to navigate in a large variety of environments.

%Tsounis \emph{et al.} \cite{Tsounis2020arxiv_deepgait} introduce a two-level hierarchical locomotion controller which plans motions via a learned Gait Planner and executes them using a learned Gait Controller. This controller learns to carefully plan footsteps as to navigate rugged terrain with crevasses. Peng \emph{et. al.} \cite{Peng18acm_deepmimic} use imitation learning to train a locomotion policy which satisfies a user-provided goal in the style of the provided motion clips. By combining motion imitation with a heightmap of the surrounding environment, the agent learns to navigate over stairs and rugged terrain. 

% \section{PROBLEM STATEMENT} \label{section:problemstatement}
% \input{sections/problemstatement}

\section{METHODS}
In this work, we frame legged locomotion as a multi-task reinforcement learning problem (MTRL) and define each task as a type of terrain that the legged robot (agent) must traverse. To learn generalizable locomotion policies, our learning system consists of a procedural terrain generator that can efficiently generate diverse training environments, and an end-to-end visual-locomotion policy architecture that directly maps the robot's exteroceptive and proprioceptive observations to motor commands.

\subsection{Multi-Task Reinforcement Learning Formulation}
Given a distribution of tasks $\mathcal{M}$, each task $M_i \in \mathcal{M}$ is a Partially Observable Markov Decision Process (POMDP). A POMDP is tuple, $M_i = \langle \mathcal{S}, \mathcal{O}, \mathcal{A}, \mathcal{T}_i, \mathcal{R} \rangle$, where $\mathcal{S}$ is the state space, $\mathcal{O}$ is the observation space, $\mathcal{A}$ is the action space, $\mathcal{T}_i: \mathcal{S} \times \mathcal{A} \times{S} \rightarrow \mathbb{R}_+$ is the transition probability function, and $\mathcal{R}: \mathcal{S} \times \mathcal{A} \rightarrow \mathbb{R}$ is the reward function. During training, the agent is presented with randomly sampled tasks $M_i \in \mathcal{M}$ (Section \ref{ssec:task_generation}). The solution of the multi-task POMDP is a stochastic policy $\pi: \mathcal{O} \times \mathcal{A} \rightarrow \mathbb{R}_+$ that maximizes the expected accumulated reward over the episode length $T$.
\begin{displaymath}
\pi^*=\arg\max_\pi \mathop{\mathbb{E}}_{M_i \in \mathcal{M}}\left[\sum_{t=0}^T r(\boldsymbol{s}_t, \boldsymbol{a}_t)\right]
\end{displaymath}

Our problem is partially observable because of the limited sensors onboard the robot\footnote{Although we use a simulated robot due to limited access to the physical robot during COVID-19, we strive to make the simulation, including the sensor measurement, as faithful as possible to the real robot.}. The robot is equipped with a LiDAR sensor to perceive the distances $\boldsymbol{d}$ to the surrounding environment. Proprioceptive information comes from a simulated IMU sensor, which includes measurement of the roll $\phi$, pitch $\theta$, and the angular velocity of the torso $^{\beta} \bm{\omega} = (\dot{\phi}, \dot{\theta}, \dot{\psi})$, and from motor encoders that measure the robot's 12 joint angles $\boldsymbol{q}$. The complete observation at timestep $t$ is
$$\boldsymbol{s}_t = [\boldsymbol{a}^T_{t-1}, \boldsymbol{o}_t, \boldsymbol{s}_{\text{TG}}^T, g_{d, t}, g_{h, t}],$$
where $\boldsymbol{o}_t = [ \boldsymbol{d}_t^T,  ^{\beta} \bm{\omega}_t^{T}, \boldsymbol{q}_t^T, \phi_t, \theta_t,]$ are the sensor observations, $g_d$ and $g_h$ are the distance and relative heading to the target, $\bf{a}_{t-1}$ is the action at the last timestep, and $\boldsymbol{s}_\text{TG}^T$ are the parameters of the trajectory generator (Section \ref{ssection:visual_locomotion_policy}). Unlike some prior work in MTRL, where the task ID is part of the observation \cite{Yu19arxiv_meta_world, Yu19neurips_mtrl_wo_inference}, we purposefully choose not to leverage such information, because identifying tasks automatically in the real world is challenging. Instead, we would like to train a policy that can rely on its own perception input and demonstrates zero-shot generalization to new tasks, without knowing the task ID explicitly. In section \ref{sec:experimental_results}, we demonstrate our perception is crucial in learning policies which generalize well to new tasks. The output action $\bf{a}_t$ of the policy specifies the desired joint angles, which are tracked by PD controllers by the simulated robot.

We employ a simple reward function, which encourages the agent to navigate to a target location $\boldsymbol{g} = (x_g, y_g, z_g)$ (the red ball in Figure \ref{fig:eval_envs}): $$r_t = \frac{g_{d, t} - g_{d, t-1}}{\Delta t},$$where $g_{d, t}$ is the Euclidean distance from the robot to the target location at timestep $t$, and $\Delta t$ is the timestep duration. This reward can be interpreted as the speed that the robot is moving towards the target location. Once the robot's center of mass is within a threshold distance to the target location, the task is complete. 

%We utilize a common state space $\mathcal{S}$ and action space $\mathcal{A}$ such that we can use the same stochastic policy $\pi_\theta: \mathcal{S} \times {A} \rightarrow \mathbb{R}_+$ for all tasks.

\begin{table}[!t]
  \vspace{1.3em}
  \caption{Terrain parameterization and generation for selected examples.}
  \label{terrain-parameterization}
  \centering
  \begin{tabular}{c ? m{3.0cm} | m{3.2cm}}
    \multirow{2}{*}{\shortstack{\textbf{Terrain}}} &\multicolumn{2}{c}{\textbf{Terrain Parameterization}} \\ 
                                & Parameters $\boldsymbol{\phi}$             &Terrain Generation        \\
     \Xhline{3\arrayrulewidth} 
    %  \midrule
     Flat                       & No parameters           & $\mathbf{H} = \mathbf{0}$   \\
     \Xhline{1\arrayrulewidth}
     Rugged                     & \shortstack[l]{
                                    Min terrain height: $h_{\text{min}}$ \\ 
                                    Max terrain height: $h_{\text{max}}$ \\
                                    Gaussian kernel std: $\sigma$}
                                & \shortstack[l]{
                                    $\mathbf{H} \sim \mathcal{U}_{m,n}(h_\text{min}, h_\text{max})$ \\
                                    Apply Gaussian smoothing \\
                                    with $\sigma$ on $\mathbf{H}$}   \\
     \Xhline{1\arrayrulewidth}
     Holes                      & \shortstack[l]{
                                    Number of holes: $n$ \\
                                    Hole depth: $h$}
                                & \shortstack[l]{
                                    $\mathbf{H} = \mathbf{0}$ \\
                                    Sample $n$ index pairs $(i, j)$  \\
                                    $\mathbf{H}(i, j) = h$} \\
     \Xhline{1\arrayrulewidth}
     Obstacles                  & \shortstack[l]{
                                    Number of obstacles: $n$ \\
                                    obstacle height: $h$}
                                & \shortstack[l]{
                                    $\mathbf{H} = \mathbf{0}$ \\
                                    Sample $n$ index pairs $(i, j)$\\
                                    $\mathbf{H}(i, j) = h$}    \\
     \Xhline{1\arrayrulewidth}
     Stairs                     & \shortstack[l]{
                                    Stair step height: $h$ \\ 
                                    Stair step length: $l$}
                                & \shortstack[l]{
                                    $\mathbf{H}(0, :)=\mathbf{0}$ \\
                                    Set column lengths to $l$\\
                                    $\mathbf{H}(i+1, :)=\mathbf{H}(i, :)+h$}\\
  \end{tabular}
\end{table} % Terrain Parameterization

\subsection{Terrain Parameterization and Procedural Task Generation} \label{ssec:task_generation}

We develop a procedural terrain generator to generate diverse and challenging terrains that provide the robot with a large quantity of rich training data. The environment is composed of $m \times n$ pillars, each pillar having cross-sectional dimensions of $l, w $, and height $h$. We denote $\mathbf{H}=\{h_{i,j}
\}\in \mathbb{R}_{m\times n}$ as the height field for all the pillars. During training, we select a task $M_i$ and adjust each pillar's heights to reflect the chosen task. Each task is a set of randomly generated terrains that belongs to the same type (e.g., flat, stairs). Each type of terrain is described by a parameter vector $\boldsymbol{\phi}_i$, which provides the lower and upper bounds for the random sampling. The terrain generator constructs the heightfield $\mathbf{H}$ from the given parameter vector $\boldsymbol{\phi}$. For example, the parameter vector $\boldsymbol{\phi}$ for the rugged terrain task (Fig. \ref{fig:rugged}) includes the minimum and maximum values of the heightfield; for the stairs task, the parameter vector defines the height and length of each step. Table \ref{terrain-parameterization} summarizes the parameters and terrain generator for selected terrain types. With this simple parameterization, we can generate over ten different types of terrains that a robot may encounter in the real world. Our procedural terrain generation algorithm provides a rich set of training data essential for generalizable policies to emerge.

\begin{figure}[!b]%
\centering
\subfloat[Visual-locomotion policy architecture.]{%
\label{fig:learned_policy}%
\includegraphics[width=0.8\linewidth]{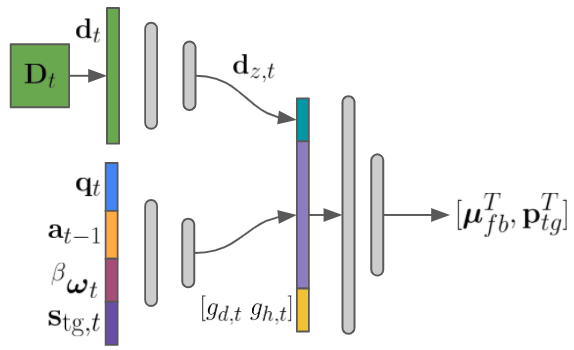}}%
\hfill
\subfloat[The locomotion component using PMTG \cite{Iscen18pmlr_PMTG} for smooth and cyclic actuation patterns.]{%
\label{fig:pmtg_diagram}%
\includegraphics[width=0.8\linewidth]{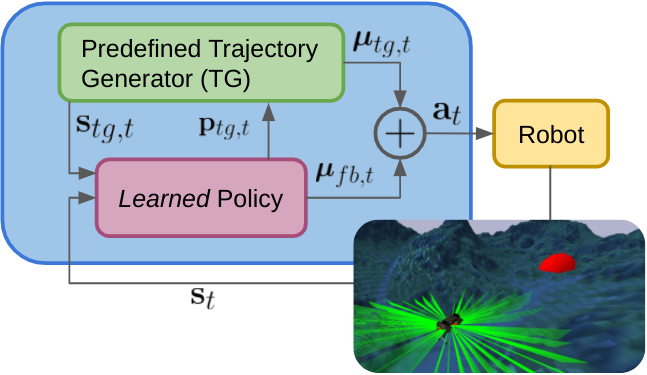}}%
\caption{Overview of the visual-locomotion policy architecture.}
\label{fig:pmtg}
\end{figure} % PMTG Policy architecture.
\begin{figure*}[!t]
\centering
\subfloat[Obstacles]{\includegraphics[width=0.24\linewidth]{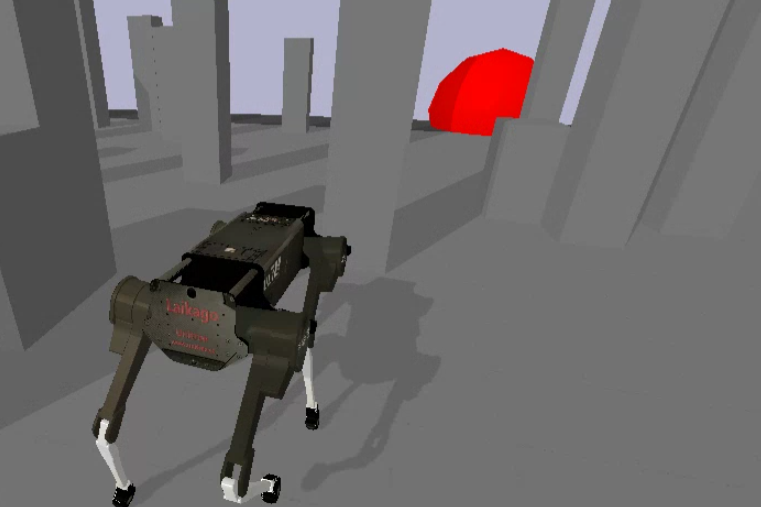}%
\label{fig:obstacles}}
\hfil
\subfloat[Rugged]{\includegraphics[width=0.24\linewidth]{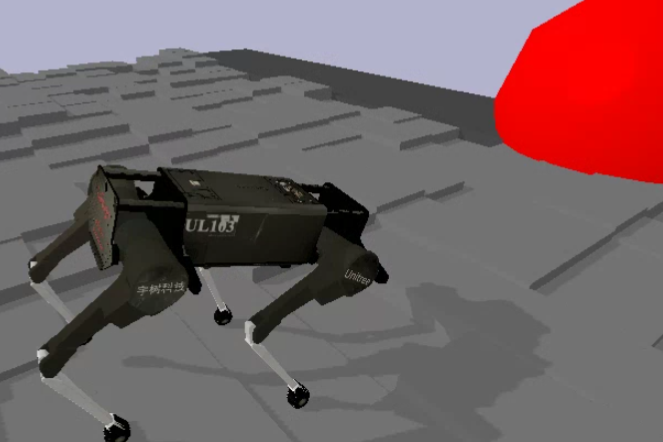}%
\label{fig:rugged}}
\hfil
\subfloat[Stairs]{\includegraphics[width=0.24\linewidth]{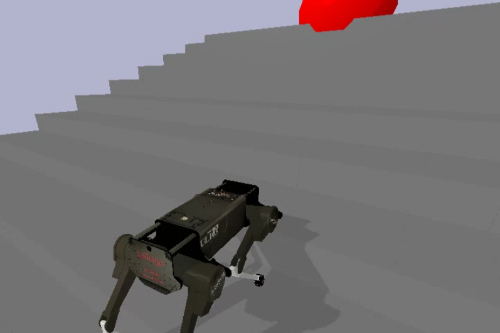}%
\label{fig:stairs}}
\hfil
\subfloat[Cliff]{\includegraphics[width=0.24\linewidth]{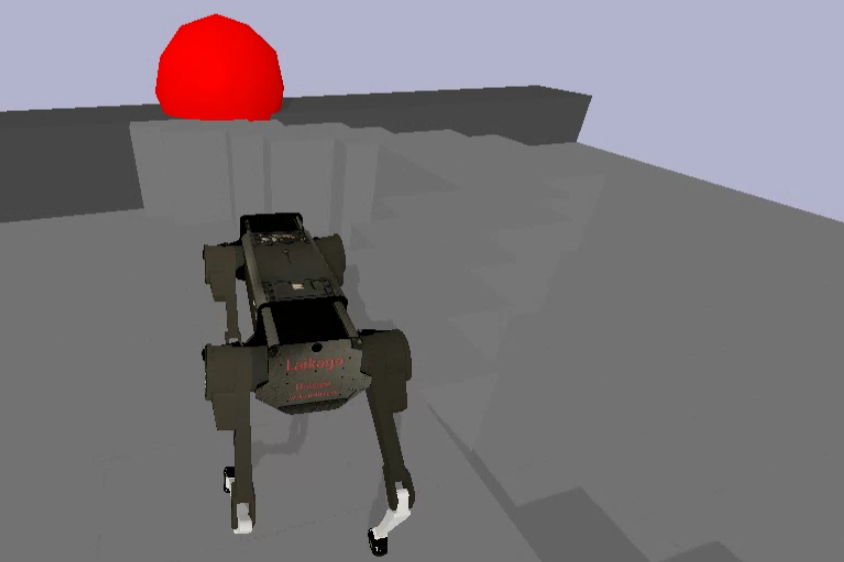}%
\label{fig:cliff}}
\hfil
\subfloat[Forest]{\includegraphics[width=0.24\linewidth]{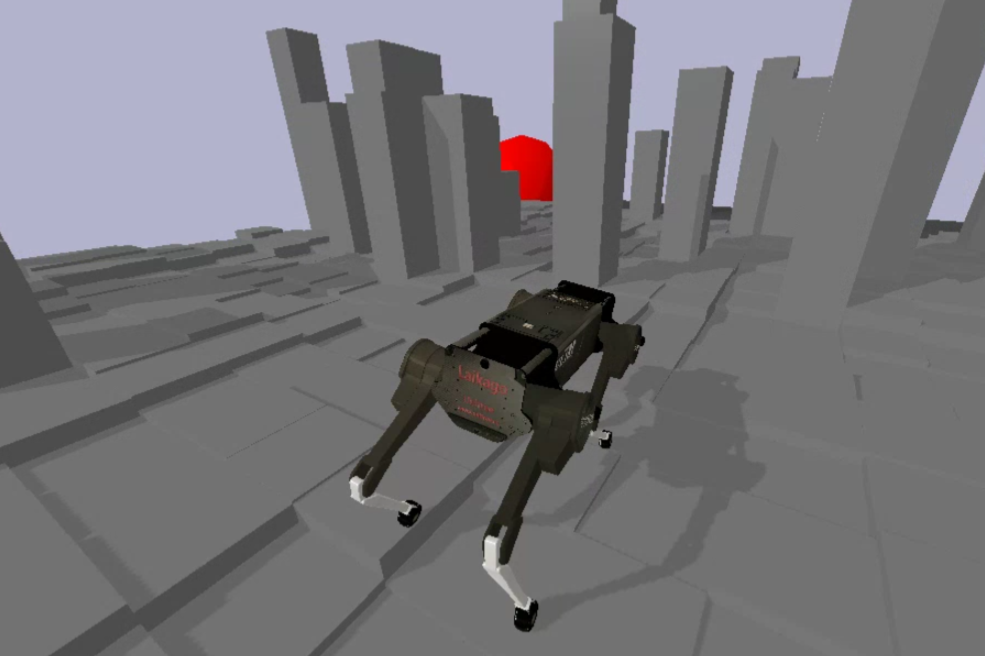}%
\label{fig:forest}}
\hfil
\subfloat[Holes]{\includegraphics[width=0.24\linewidth]{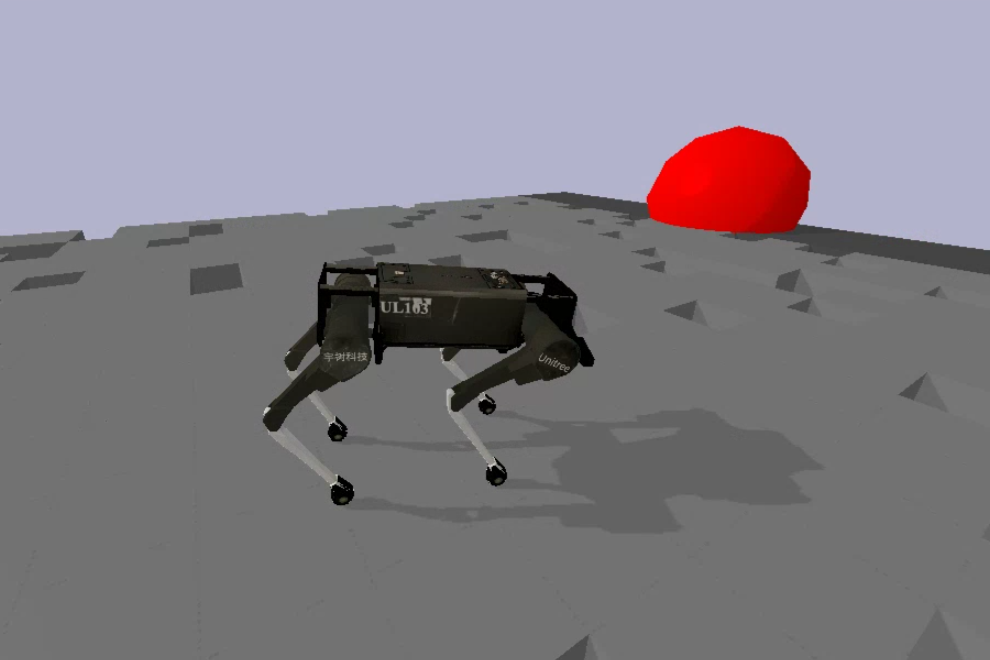}%
\label{fig:holes}}
\hfil
\subfloat[Gaps]{\includegraphics[width=0.24\linewidth]{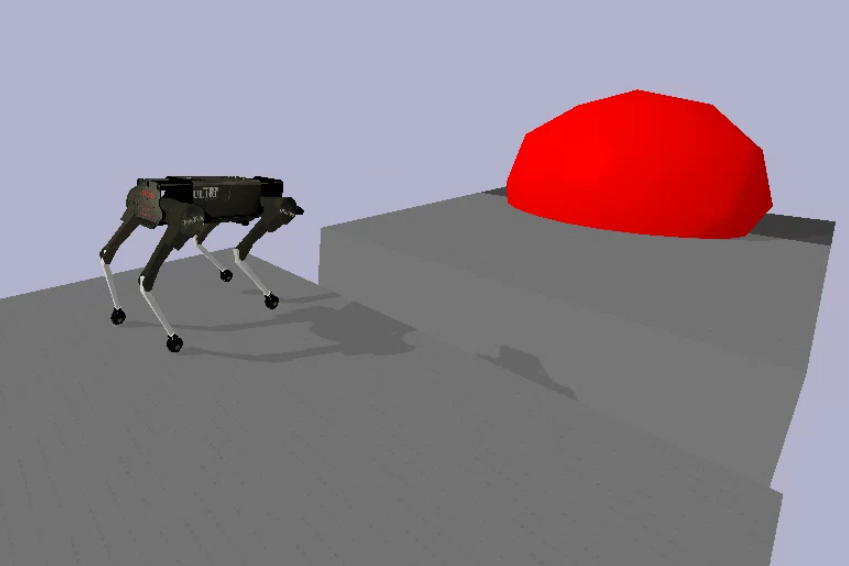}%
\label{fig:gap}}
\hfil
\subfloat[Hills]{\includegraphics[width=0.24\linewidth]{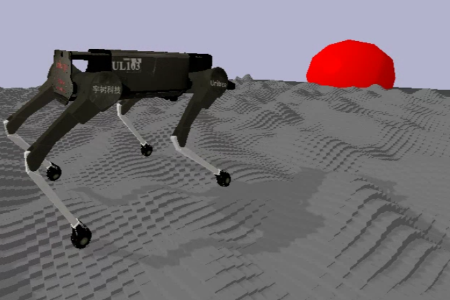}%
\label{fig:hills}}
\hfil
\caption{A Laikago robot deployed in various procedurally generated training environments. The red sphere represents goal $g$ and success radius $r_g$.}
\label{fig:pillars}
\end{figure*} % Train environments.

\subsection{Visual-Locomotion Policy Architecture} \label{ssection:visual_locomotion_policy}
Exteroceptive perception plays a crucial role when legged robots need to navigate different terrains and environments with obstacles and humans \cite{Matthis18CurrBio, Matthis14ExpPsych}. As such, we aim to incorporate perception into our policy architecture such that information from the robot's surroundings can modulate locomotion. Additionally, the policy's low-level actuation commands need to be smooth and realizable on the physical robot. To this end, we seek to restrict the search space of possible gaits to be cyclic and smooth while still expressive enough so that the perception can modulate locomotion sufficiently to work on different terrains.

In our visual-locomotion policy architecture (Fig. \ref{fig:pmtg}), we use two separate neural network encoders to process the proprioceptive and exteroceptive inputs. The upper branch of Fig. \ref{fig:learned_policy} processes the LiDAR input, while the lower branch takes care of proprioceptive information. The learned lower-dimensional features are concatenated with the target information before being passed to the policy's locomotion component. We chose to use Policies Modulating Trajectory Generators (PMTG) \cite{Iscen18pmlr_PMTG} as our locomotion component architecture (Fig. \ref{fig:pmtg_diagram}). PMTG encourages the policy to learn smooth and cyclic locomotion behaviors. PMTG outputs a desired trajectory for the legs that is modulated by a learned policy $\pi_\theta(\cdot)$: The policy observes the state of the trajectory generator (TG), $\boldsymbol{s}_{\text{tg}}$, and the robot's observation $\boldsymbol{s}_t$, then outputs parameters of the TG, $\boldsymbol{p}_{\text{tg}}$, including gait frequency, swing height, and stride length, and a residual action term $\boldsymbol{\mu}_{fb}$. The final output action of our visual-locomotion policy is the combination of the trajectory generator and the residual action: $\boldsymbol{a}_t = \boldsymbol{\mu}_{tg} + \boldsymbol{\mu}_{fb}$. Please refer to the original paper \cite{Iscen18pmlr_PMTG} for more details. As detailed in \cite{Hess17arxiv_locomotion_emergence}, our visual-locomotion policy architecture achieves a separation of concerns between the basic locomotion skills and terrain perception, which enables the robot to adapt its smooth locomotion behaviors according to its surrounding environments.

\section{EXPERIMENTAL RESULTS} \label{sec:experimental_results}

We design experiments to validate the proposed system's ability to learn a visual locomotion policy that generalizes well to terrains not encountered during training. In particular, we would like to answer the following two questions:
\begin{itemize}
    \item Can our system learn visual locomotion policies that demonstrate zero-shot generalization to new terrains?
    \item Can our policy architecture effectively use LiDAR input and PMTG parameterization to improve the generalization performance over unseen terrains?
\end{itemize}

\subsection{Experiment Details}
To answer the above questions, we evaluate our system using a simulated Unitree Laikago quadruped robot \cite{laikago2018}, which weighs approximately 22kg and is actuated by 12 motors. We simulate the onboard Velodyne VLP-16 (Puck) LiDAR sensor, which provides the perception of the surrounding environment (See Figure \ref{fig:pmtg_diagram}). The LiDAR measures the distance from the surrounding obstacles and terrain to the robot. This sensor supports 16 channels, a 360$^\circ$ horizontal field of view, and a 30$^\circ$ vertical field of view. We add Gaussian noise to the ground-truth distance readings in simulation to mimic the real-world noise model. The 3D LiDAR scan matrix $\bf{D}$ is normalized to range $[0, 1]$ and flattened to a vector $\bf{d}$. 

Our policy computes joint target positions ($\bm{a}_t$), which are converted to target joint torques by a PD controller running at $1\si{\kilo\hertz}$.
Rigid body dynamics and contacts are also simulated at $1\si{\kilo\hertz}$.
In other words, the position and velocity (provided by PyBullet \cite{coumans2019}) and the desired torque (provided by the PD controller) are sent to the actuator model every $1\si{\ms}$. The actuator model then computes $10$ internal $100\si{\us}$ steps and provides the effective output torque of the actuator, which is then used by PyBullet to compute joint accelerations.
The simulation environment is configured to use an \emph{action repeat} of $10$ steps, which means that our policy computes a new action $\bm{a}_t$ and receive a state $\bm{s}_t$ every $10\si{\ms}$ ($100\si{\hertz}$).

We train the visual-locomotion policy using the MTRL formulation with simulated environments randomly generated using our procedural task generation method (Section \ref{ssec:task_generation}). We choose a distributed version of the Proximal Policy Optimization (PPO) \cite{schulman2017proximal} in TF-Agents \cite{TFAgents} for training. We use a 2-layer fully-connected neural network of dimensions $(512, 256)$ to parameterize the value function and another network of dimensions $(256, 128)$ to parameterize the policy. The policy outputs the parameters of a multivariate Gaussian distribution, which we sample actions from during training. We use a greedy policy during evaluation by executing the mean of the multivariate Gaussian distribution provided by the policy network. The dimensions of the exteroceptive and proprioceptive input encoders are both $(32, 16,4)$, respectively. We use the ReLU activation function for all layers in both networks \cite{andrychowicz2020matters}. The advantages are estimated using Generalized Advantage Estimation \cite{schulman2018highdimensional}.

We then evaluate the trained policies on a suite of testing environments not encountered during training. Figure \ref{fig:eval_envs} illustrates a subset of these testing environments. These high-fidelity simulated environments are created in PyBullet physics engine \cite{coumans2019} with Gibson scenes \cite{fei18cvpr_gibson}. A policy's ability to successfully navigate across a given terrain is measured using the \emph{task completion rate}, $tcr$, which measures how close the agent gets to the target relative to its starting position:
$$tcr = 1 - \frac{g_{d,T}}{g_{d,0}},$$
where $g_{d,T}$ is the final Euclidean distance between the robot and the target when the robot falls or completes the task, and $g_{d,0}$ is the distance at the beginning of the episode. A task completion rate of 1 indicates successful navigation to the target, whereas $tcr$ close to zero means that the robot cannot navigate across the terrain.

\begin{figure}[!b]
\centering
\includegraphics[width=1.0\linewidth]{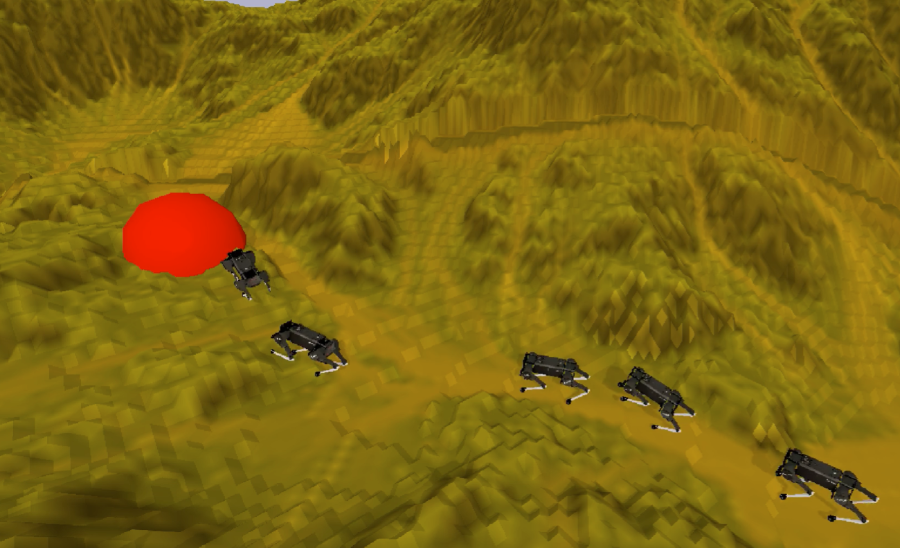}
% \subfloat[Flat]{\includegraphics[width=0.5\linewidth]{Images/Return Curves/flat.png}%
% \label{fig:flat_curve}}
\caption{ Snapshot of a laikago robot navigating through mountainous terrain not encountered during training. Please refer to the supplementary video for more examples of the agent navigating challenging terrains. }
\label{fig:mountain_traj}
\end{figure} % Return curves.

\begin{table*}[!h]
  \vspace{1.3em}
  \caption{Generalization performance of our visual-locomotion policy.}
  \label{eval-table-1}
  \centering
  \includegraphics[width=0.83\linewidth]{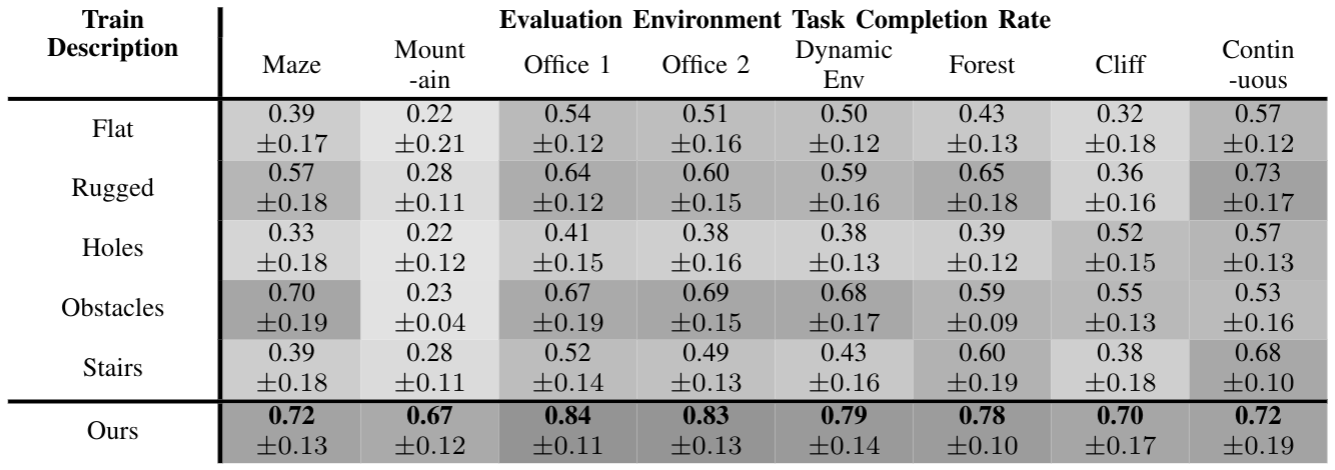}
\end{table*} % Eval table train task.

\begin{figure}[!t]
\captionsetup[subfloat]{farskip=0pt,captionskip=0pt}
\centering
\subfloat{\includegraphics[width=0.88\linewidth]{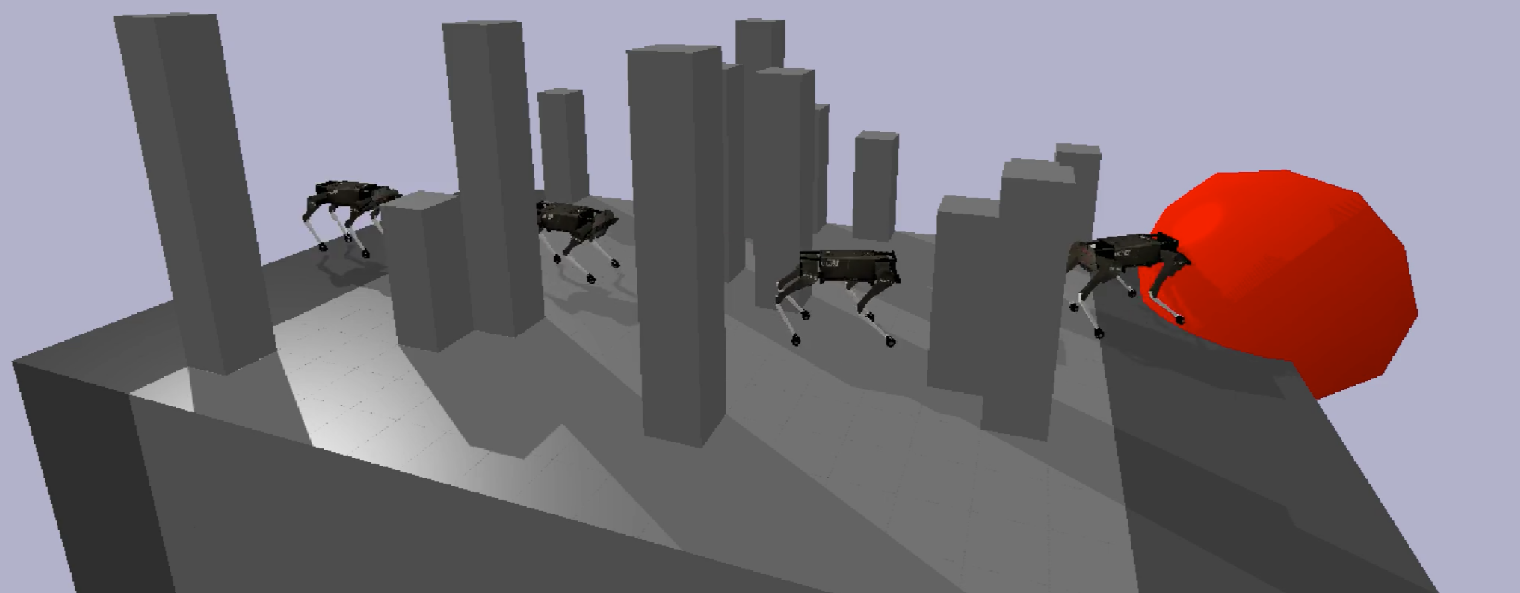}%
\label{fig:obstacle_blend}}
\\[-0ex]
% \hfil
\subfloat{\includegraphics[width=0.815\linewidth]{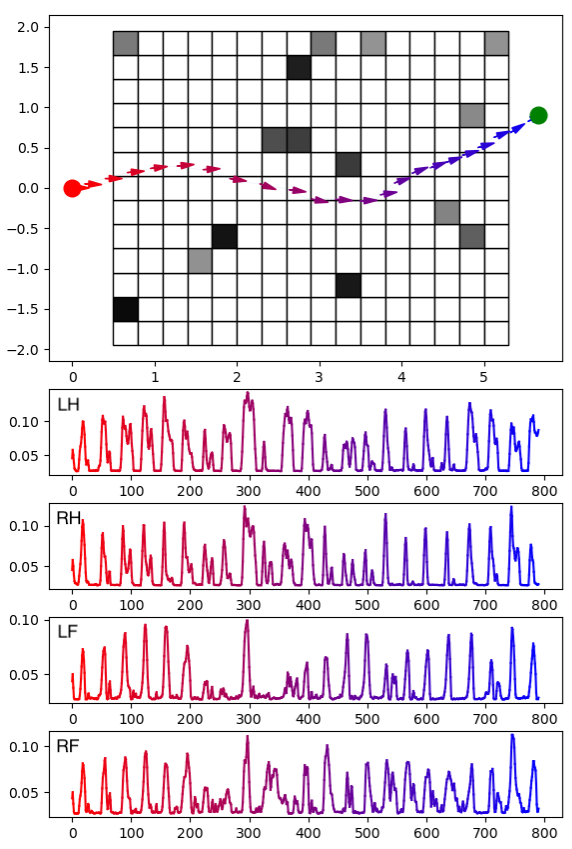}%
\label{fig:obstacle_traj}
}

\caption{ Visualization of trajectory generated by our method in an environment with many obstacles. Foot Z positions for the left hind, right hind, left forward, and right forward feet are shown. }
\label{fig:obstacle_traj_fig}
\end{figure} % Obstacle Blend
\begin{figure}[!t]
\captionsetup[subfloat]{farskip=0pt,captionskip=0pt}
\centering
\subfloat{\includegraphics[width=0.9\linewidth]{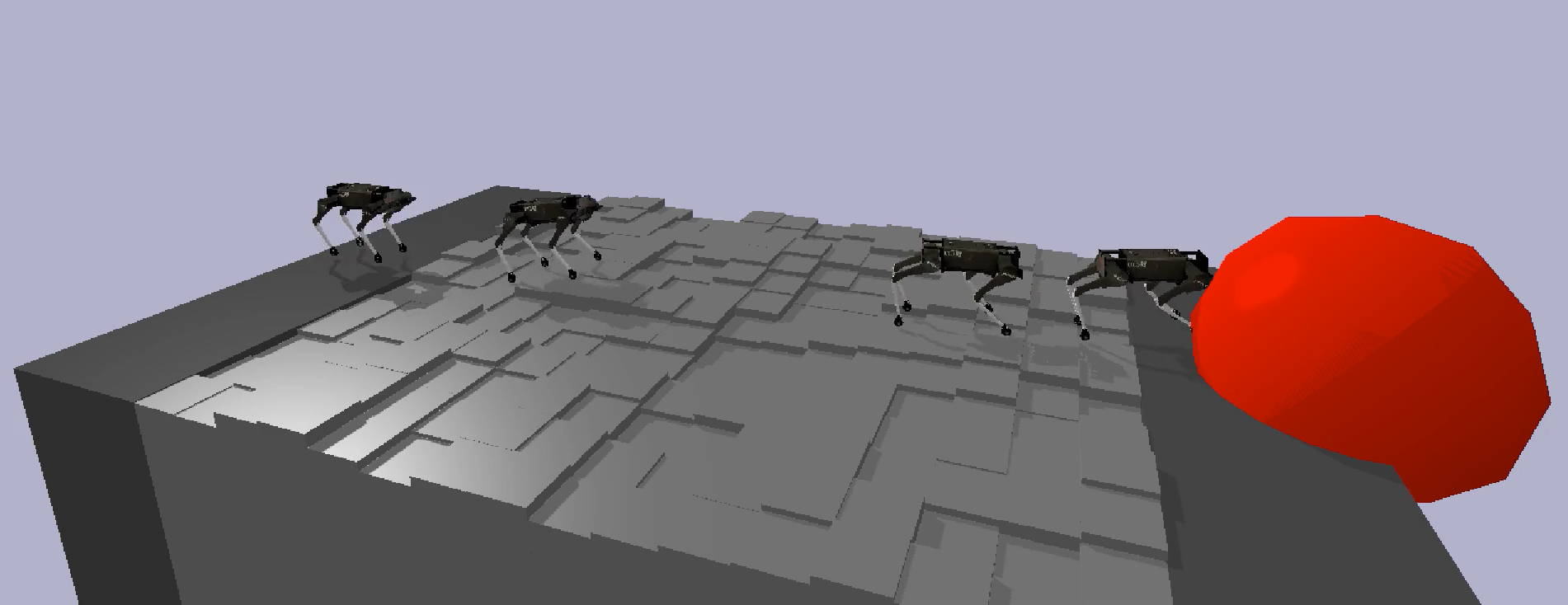}%
\label{fig:rugged_blend}}
\\[-0ex]
% \hfil
\subfloat{\includegraphics[width=0.85\linewidth]{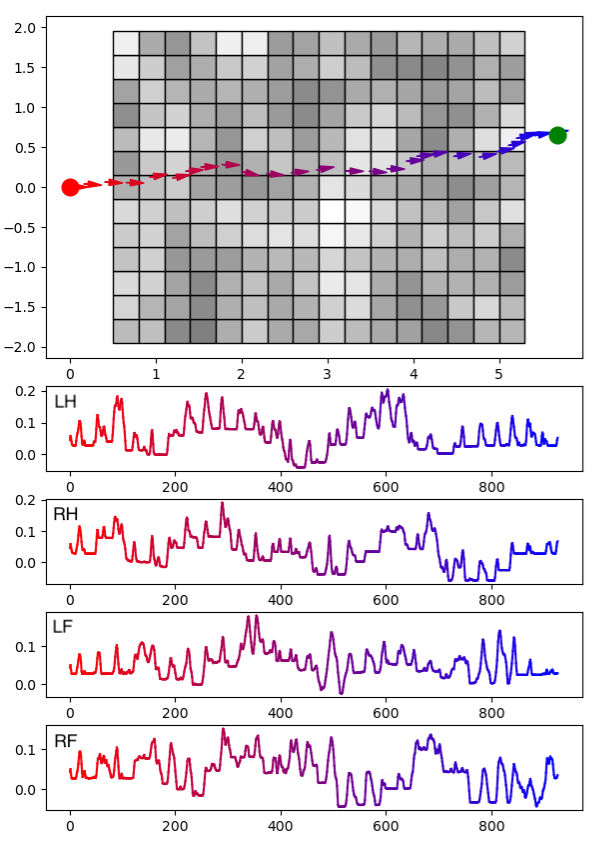} %
\label{fig:rugged_traj}
}

\caption{ Visualization of trajectory generated by our method in a rugged terrain. Foot Z positions for the left hind, right hind, left forward, and right forward feet are shown. The rugged terrain requires that the robot carefully place its feet to maintain balance.}
\label{fig:ruggged_traj_fig}
\end{figure} % Rugged Blend

\begin{table*}[!t]
  \caption{Comparison of our proposed method to other policies deployed in a MTRL training regime. The performance decreases when the policy does not use a PMTG parameterization, when the policy is not provided exteroceptive inputs from the LiDAR, and when multi-task training is performed in a sequential manner.}
  \label{eval-table-2}
  \centering
  \includegraphics[width=0.86\linewidth]{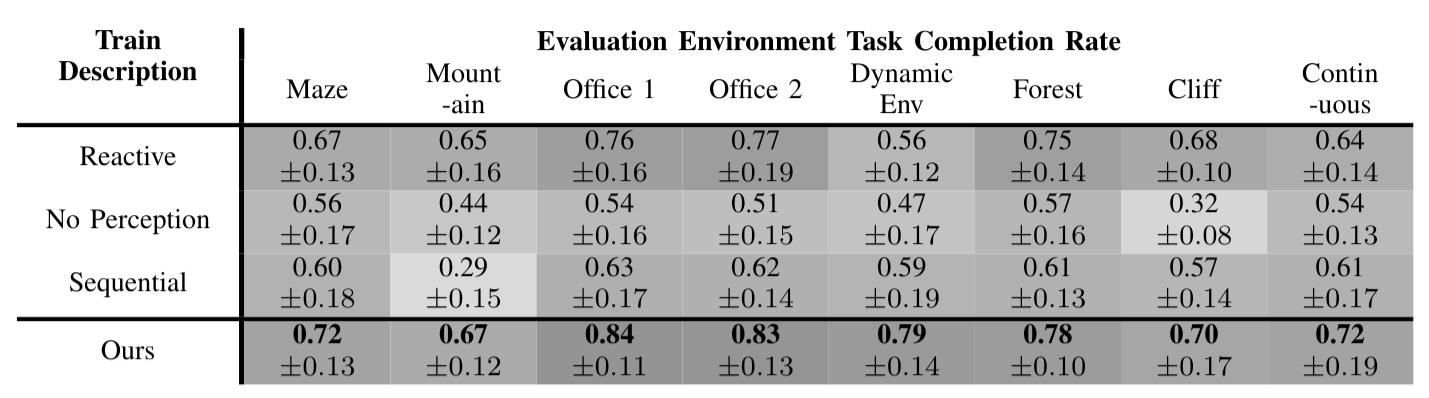}
\end{table*} % Eval table ablation.

\subsection{The Impact of MTRL on Generalization}
Table \ref{eval-table-1} shows the generalization performance of our visual-locomotion policy trained on different types of terrains (rows) and tested in unseen environments (columns), including a maze (Maze), a steep and rugged mountain (Mountain), two indoor scenarios (Office 1 and Office 2), an office space with moving humans (Dynamic Env), a forest scene with rugged terrain and obstacles (Forest), a winding path with a cliff on both sides (Cliff), and a randomly-generated continuous mesh (Continuous). Policies trained on a single type of terrain achieve a low task completion rate in the testing environments due to a lack of diverse training data. In contrast, our approach achieves much higher generalization performance. For instance, our method on average achieves a task completion rate of $67\%$ on the mountain task, while policies trained in a single type of terrain only achieve $28\%$ at best (See Figure \ref{fig:mountain_traj} for a snapshot of our policy navigating up the rugged mountain trail). These results indicate that our MTRL formulation using procedural task generation, and visual-locomotion policy architecture, results in superior generalization performance. The policy learned with our system can be successfully deployed in new unseen environments.

\subsection{Ablation Studies}
We perform three ablation studies to understand the importance of each design decision in our system. Table \ref{eval-table-2} summarizes their impacts on the resulting generalization performance of the policy.

\paragraph{PMTG} We replace the locomotion component of the visual-locomotion policy with a reactive policy that does not have a trajectory generator. Our PMTG-parameterized visual-locomotion policy performs 28\%-218\% better than a pure reactive locomotion component. We find that PMTG produces smoother actions and leads to improved zero-shot generalization to new terrains.

\paragraph{Exteroceptive input} We remove the LiDAR input from the visual-locomotion policy. Observing Table \ref{eval-table-2}, it is clear that the exteroceptive information plays a critical role in learning generalizable locomotion policies that can adapt to a wide variety of terrains. This finding agrees with results from the field of experimental psychology, which establish the importance of exteroceptive observations in guiding foot placement when navigating over complex terrains \cite{Matthis14ExpPsych, Matthis18CurrBio}. Figure \ref{fig:obstacle_traj_fig} visualizes the trajectory produced by our visual locomotion policy in a terrain with obstacles. When walking over flat terrain, the robot's foot height is constant and cyclic, only varying when turning to avoid obstacles. In contrast, on rugged terrain (Figure \ref{fig:ruggged_traj_fig}), the robot carefully places its feet to adapt to the geometry of the terrain to maintain balance. This careful foot placement is essential for challenging terrains and requires a visual feedback loop, which our learning system can provide.

\paragraph{MTRL training scheme} Our system generates a new random locomotion task at each episode for all the distributed workers. This ensures a steady stream of rich training data to the agent. In this study, we lower the variety of tasks supplied in a single training step by proving tasks sequentially. That is, the agent learns one task for a fixed number of training steps before switching the task. The policy trained in a sequential fashion performs poorly due to \emph{catastrophic forgetting} \cite{McCloskey89plm_forgetting}.

These ablation studies confirm the importance of each component of our system, including the exteroceptive input and PMTG used in the visual-locomotion policy architecture, as well as our multi-task POMDP training formulation. By combining these components, our system can learn locomotion policies that work on various terrains and demonstrate zero-shot generalization to new environments.

\section{CONCLUSION}
We introduce a learning system that enables legged robots to traverse various environments and demonstrates zero-shot generalization to new terrains. Our system consists of a novel multi-task reinforcement learning formulation of the locomotion problem, a visual locomotion policy architecture that encourages smooth actions and incorporates perception to modulate locomotion, and a novel procedural terrain generation algorithm that provides the agent with rich training data from a variety of simulated terrains. Our results on a suite of simulated environments show that treating legged locomotion as a multi-task POMDP leads to increased generalization performance. Additionally, we show that providing the policy with a strong prior over the space of gaits further enhances its ability to generalize to unseen terrains. In future work, we plan to evaluate our work on a real-world robot.

% \section*{ACKNOWLEDGMENT}

% The preferred spelling of the word “acknowledgment” in America is without an “e” after the “g”. Avoid the stilted expression, “One of us (R. B. G.) thanks . . .”  Instead, try “R. B. G. thanks”. Put sponsor acknowledgments in the unnumbered footnote on the first page.

\clearpage

%%%%%%%%%%%%%%%%%%%%%%%%%%%%%%%%%%%%%%%%%%%%%%%%%%%%%%%%%%%%%%%%%%%%%%%%%%%%%%%%
\bibliography{references}  % .bib
\bibliographystyle{IEEEtran}

\end{document}